\documentclass[10pt,twocolumn,letterpaper]{article}

\usepackage[pagenumbers]{cvpr} 

\definecolor{cvprblue}{rgb}{0.21,0.49,0.74}
\usepackage[pagebackref,breaklinks,colorlinks,allcolors=cvprblue]{hyperref}

\usepackage[framemethod=tikz]{mdframed}
\usepackage{amsmath}
\usepackage{amsfonts} 
\usepackage{latexsym}
\usepackage{graphicx}
\usepackage{stackengine}
\usepackage{booktabs}
\usepackage[utf8]{inputenc}
\usepackage{xcolor}
\usepackage{colortbl}
\usepackage{multirow}
\usepackage{enumitem}
\usepackage{algorithm}
\usepackage{algpseudocode}
\usepackage[accsupp]{axessibility}

\usepackage{amsmath}
\usepackage{calc}
\usepackage{cases}
\newcommand{\ourmethod}{\texttt{DyFo}\xspace}
\newcommand{\ourmethodlong}{\texttt{Dynamic Focus}\xspace}
\newcommand{\focus}{\textit{focus}\xspace}

\def\paperID{2715} 
\def\confName{CVPR}
\def\confYear{2025}

\title{DyFo: A Training-Free Dynamic Focus Visual Search for Enhancing LMMs in Fine-Grained Visual Understanding }

\author{
Geng Li\textsuperscript{1} \quad
Jinglin Xu\textsuperscript{2} \quad
Yunzhen Zhao\textsuperscript{3} \quad
Yuxin Peng\textsuperscript{1} \thanks{Corresponding author.} \\
\textsuperscript{1}Wangxuan Institute of Computer Technology, Peking University \\
\textsuperscript{2}School of Intelligence Science and Technology, University of Science and Technology Beijing \\
\textsuperscript{3}Tencent Beijing Research, Beijing, 100193, China \\
{\tt\small ligeng@stu.pku.edu.cn, xujinglinlove@gmail.com, yunzhenzhao@tencent.com, pengyuxin@pku.edu.cn}
}

\begin{document}

\maketitle
\begin{abstract} 

Humans can effortlessly locate desired objects in cluttered environments, relying on a cognitive mechanism known as visual search to efficiently filter out irrelevant information and focus on task-related regions. Inspired by this process, we propose \ourmethod (\ourmethodlong), a training-free dynamic focusing visual search method that enhances fine-grained visual understanding in large multimodal models (LMMs). Unlike existing approaches which require additional modules or data collection, \ourmethod leverages a bidirectional interaction between LMMs and visual experts, using a Monte Carlo Tree Search (MCTS) algorithm to simulate human-like focus adjustments. This enables LMMs to focus on key visual regions while filtering out irrelevant content, without introducing additional training caused by vocabulary expansion or the integration of specialized localization modules. Experimental results demonstrate that \ourmethod significantly improves fine-grained visual understanding and reduces hallucination issues in LMMs, achieving superior performance across both fixed and dynamic resolution models. The code is available at \url{https://github.com/PKU-ICST-MIPL/DyFo_CVPR2025} \end{abstract}
    
\section{Introduction}
\label{sec:intro}

Humans can quickly locate desired objects in a cluttered environment or recognize familiar faces in a crowd. This efficient visual information retrieval ability is attributed to a visual processing mechanism known as \textbf{visual search}, which has been extensively studied in cognitive and vision sciences~\cite{torralba2006contextual,peelen2011neural,wolfe2011visual, wolfe2020visual,wang2023statistical}.
In visual search, dynamic focusing adjustments are taken to efficiently filter out irrelevant visual information and concentrate on task-related regions \cite{RAYNER19953eyemove, rucci2015fixational}. For example, when reading, people focus on semantically relevant text regions (fixations) and quickly scan past irrelevant areas (saccades) \cite{pritchard1960visual_fixational,huey1908read}. This inspires us to recognize the importance of selecting visual inputs based on semantic relevance for efficient visual processing.

\begin{figure*}[t]
\centering
\includegraphics[width=\textwidth]{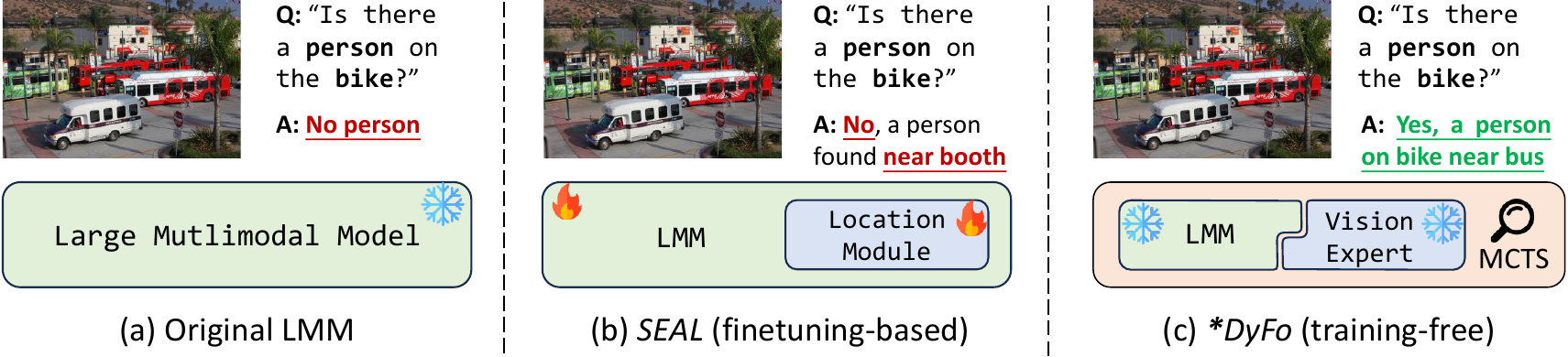}
\caption{An illustration of three mechanisms for large multimodal models (LMMs) in fine-grained visual understanding tasks.}
\label{fig:first}
\vspace{-6pt}
\end{figure*}

As a promising paradigm for achieving general multimodal perception and reasoning capabilities, LMMs have achieved notable performance improvements across a range of multimodal tasks~\cite{blip2, alayrac2022flamingo, minigpt4, llava, instructblip}. Some LMMs, such as BLIP-2\cite{blip2} and LLaVA\cite{llava}, use fixed-resolution encoders to process visual information. These encoders, like ViT\cite{vit}, operate at a specified input resolution (224 or 336), converting scaled images into a fixed number of tokens.
This approach often faces difficulties with high-resolution image tasks because downsizing images and limiting the number of tokens cause the loss of important details, such as small objects, which can eventually lead to hallucinations\cite{guan2024hallusionbench,wang2024amberllmfreemultidimensionalbenchmark}. To handle high-resolution images, some works (e.g., Qwen2-VL \cite{Qwen2VL} and LLaVA-Next \cite{li2024llavanextinterleavetacklingmultiimagevideo}) divide images into smaller sections. Each section is processed separately to generate fixed number of tokens, which are then combined. This approach supports dynamic resolutions and variable token counts, greatly enhancing LMMs' ability to process high-resolution images. However, real-world visual inputs often consist of complex scenes with many unrelated objects, such as busy streets crowded with pedestrians and vehicles or supermarket shelves filled with products. In fine-grained visual understanding tasks that require focusing on specific visual subregions to answer detailed questions, unrestrictedly increasing visual input may introduce a large amount of irrelevant content, causing interference. Some studies \cite{yu2024hallucidoctor, wang2023evaluationanalysishallucinationlarge} have shown that as irrelevant input increases, hallucinations in large multimodal models (LMMs) become more pronounced. Therefore, merely expanding resolution and input capacity may further introduce hallucinations, ultimately limiting the performance of fine-grained visual understanding tasks.

To enhance LMMs' fine-grained visual understanding~\cite{fine_grain_cite_1,fine_grain_cite_2,fine_grain_cite_3}, some studies \cite{peng2023kosmos, chen2023shikra, you2023ferret} use region bounding boxes or masks during training to enable local region understanding. However, this approach requires user-provided spatial prompts (e.g., boxes or masks) at inference, which are  unavailable in real-world applications. SEAL \cite{wu2024vstar} combines visual search with LMMs to assist in automatically locating specific visual regions to improve fine-grained understanding. To support multi-turn search, SEAL utilizes a visual working memory to store intermediate results and fine-tune the system to adapt. Besides, it requires modifications to the LMM's vocabulary and additional localization modules, such as object detection and heatmap generation, to enable visual search. While SEAL achieves strong results, the need for extensive data collection and the adaptation of additional modules limits its applicability to general-purpose LMMs like Qwen2-VL. Moreover, recent study \cite{zhang2024psalmpixelwisesegmentationlarge, glamm,you2023ferret,chen2023shikra} show that methods incorporating extra modules with LMM to support ground task still lag behind visual expert models (e.g., Grounding Dino \cite{groundingdino}, Lang SAM \cite{langsam}) in visual task performance, hindering further improvements in visual search.

Inspired by human dynamic focusing adjustments, we propose \ourmethod (\ourmethodlong), a training-free visual search method that enables LMMs to incorporate the abilities of visual experts, selectively amplifying critical visual information and filtering irrelevant input to enhance fine-grained visual understanding tasks. \ourmethod leverages bidirectional interaction between LMMs and visual experts to directly focus on key visual regions, and through Monte Carlo Tree Search (MCTS) \cite{silver2016mastering}, simulates human-like focus adjustments in visual search. Unlike existing methods like SEAL, \ourmethod does not require vocabulary expansion or additional dependencies on instruction-tuned localization or heatmap modules. With a consensus-based reward mechanism between the LMM and the visual expert and an action space that simulates human focusing adjustments, \ourmethod provides an “out-of-the-box” dynamic focusing visual search solution. Experimental results demonstrate that our method effectively enhances fine-grained visual understanding and broadly reduces visual hallucination issues. In summary, our contributions are threefold:
\begin{itemize}
    \item We propose \ourmethod, a training-free visual search method that significantly improves fine-grained visual understanding tasks for both fixed and dynamic resolution-supported models.
    \item We introduce an MCTS-based collaboration paradigm, allowing the use of high-performance visual experts without costly LMM training or architectural changes.
    \item We validate the superior performance of our method in general hallucination and fine-grained visual understanding tasks, demonstrating its potential to enhance visual capabilities.
\end{itemize}

\section{Related work}
\label{sec:related_work}

\noindent \textbf{Large Multimodal Models.}
Since the advent of large language models (LLMs), their success in diverse language applications has led to the development of LMMs that integrate vision and language. Initially, LMMs functioned as dispatchers connecting language models with vision experts, as seen in Visual ChatGPT~\cite{visualChatGPT}, HuggingGPT~\cite{shen2024hugginggpt}, and MM-REACT~\cite{yang2023mm}. More recently, LMMs have focused on aligning visual and language modalities through training on image-caption pairs or visual question-answering datasets. Methods like LLaVA~\cite{llava} map image tokens to pre-trained LLM representations via a learned projector, while BLIP-2~\cite{li2022blip,blip2} employs a query transformer to generate image embeddings from learned queries after extracting image features. MoVA~\cite{zong2024mova} uses an adaptive router to integrate task-specific vision experts with a coarse-to-fine approach. Qwen2-VL~\cite{Qwen2VL} enhances high-resolution image understanding through dynamic resolution handling and flexible tokenization.

\begin{figure*}[t]
    \centering
    \includegraphics[width=\textwidth]{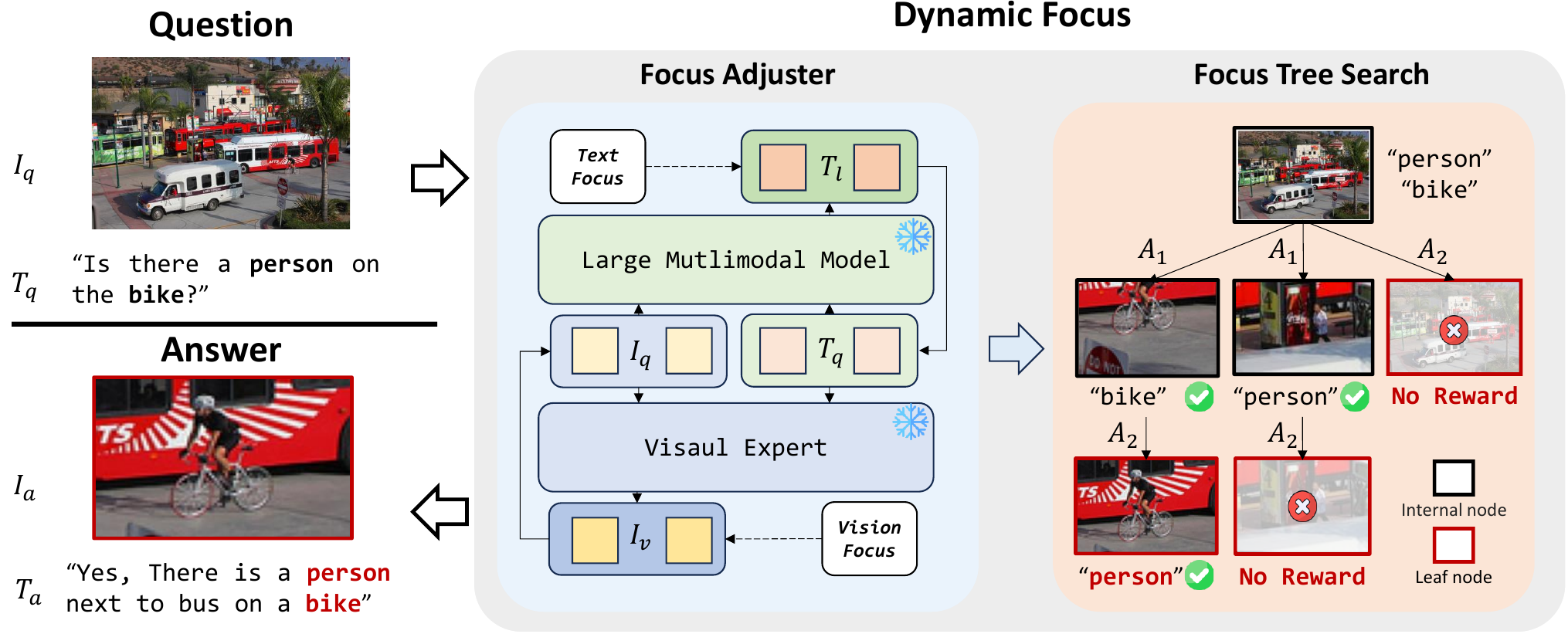}
    \vspace{-10pt}
    \caption{An illustration of \ourmethod framework, composed by \textbf{Focus Adjuster} and \textbf{Focus Tree Search} (Section~\ref{sec:our_method}).}
    \label{fig:method}
    \vspace{-12pt}
\end{figure*}

\noindent \textbf{Visual Search.} Inspired by human visual search and its connection to eye movements, early methods simulated this process using Bayesian frameworks with saliency maps \cite{sclar2020modeling, torralba2006contextual}, deep networks for similarity mapping \cite{zhang2018finding}, or inverse reinforcement learning (IRL) to learn search strategies \cite{yang2020predicting}. However, these approaches mainly replicate gaze trajectories without precise target localization and rely on fixed-size attention windows, limiting their flexibility in complex scenes.  SEAL \cite{wu2024vstar} integrates localization modules and visual memory to combine visual search with large multimodal models (LMMs) to guide attention regions, but it requires additional training for effective visual search. In contrast, our approach leverages the collaboration between LMMs and visual experts to achieve a dynamic focus mechanism without requiring extra training or module dependencies, providing a “plug-and-play” solution to enhance fine-grained visual understanding.

\noindent \textbf{Visual Experts.} Visual experts are models specifically designed to handle vision tasks with both text and image inputs, capable of generating precise image outputs guided by text. Recent studies have leveraged the ``Segment Anything Model" (SAM) for text-prompted segmentation~\citep{groundedSAM, fast_sam, refsam, lisa, lisa++}. For instance, Grounded-SAM combines text-prompted boxes generated by Grounding DINO with SAM’s segmentation function to create a two-stage framework~\citep{groundedSAM, groundingdino}. Fast-SAM matches text and image regions of interest based on CLIP feature similarity~\citep{fast_sam, clip}. RefSAM uses a cross-modal MLP to project text embeddings into SAM’s sparse and dense embeddings, while LISA leverages large multimodal models like LLaVA to extract multimodal embeddings for SAM~\citep{refsam, llava}.
\section{Method}

\subsection{Visual Search}
\label{sec:visual_search}

In our setup, a large multimodal model (LMM) with parameters $\theta$ takes an image $I \in \mathbb{R}^{H \times W \times 3}$ and a text $X = [x_1,...,x_n]$ as input, where the text contains $n$ tokens, and outputs a text $Y = [y_1,...,y_m]$ with $m$ tokens. When generating the output text $Y$, the model sample autoregressively from a probability distribution conditioned on the input $I$ and $X$. At time step $t$, the probability of token $y_t$ is
\begin{equation}
    y_t \sim p_{\theta} (y_t | I,X,y_{<t})
    \propto \exp{ \text{logit}_{\theta}(y_t | I,X,y_{<t})},
    \label{eq:prob}
\end{equation}
where $\text{logit}_{\theta}$ is the unnormalized log probability of the token  $y_t$, i.e., the value before softmax. The goal of visual search is to find the corresponding image regions $I \in \mathbb{R}^{H' \times W' \times 3}$, where $H'<H$ and $W'<W$, making \( \text{logit}_{\theta}(\hat{y}_t | I, X, y_{<t}) \) as high as possible. The $\hat{y}_t$ is the token of the right answer. Recent studies \cite{wu2024vstar} have introduced localization modules that output heatmaps and object boxes to highlight specific areas to enhance the visual grounding ability of models. However, this visual search method relies on the specific architecture of LMMs with extra localization module and introduces substantial costs in grounding data collection and LMM finetuning, hindering its transferability to other models such as Qwen2-VL. We explore a more general visual search strategy by leveraging the collaboration between existing visual experts and LMMs to achieve general and more efficacious visual search.

\subsection{Dynamic Focused Visual Search}
\label{sec:our_method}

We propose a training-free visual search method called \ourmethodlong (\ourmethod), designed to help large multimodal models (LMMs) focus on specific regions of an image through collaboration with visual experts. Specifically, we employ a Monte Carlo Tree Search (MCTS) algorithm, treating focus regions and text as nodes to simulate a human-like, focus-based search process. During this process, the LMM's textual output guides the visual expert to retrieve relevant image regions, while the visual expert’s image output, in turn, directs the LMM to adjust its textual focus, as illustrated in \cref{fig:method}.

\subsubsection{Focus Adjuster}

Although LMMs possess multimodal integration capabilities, they are often limited in fine-grained visual tasks due to constraints in resolution and hallucination caused by irrelevant content. Whether it’s fixed-resolution models (e.g., LLaVA) or dynamic-resolution models (e.g., Qwen2-VL), they struggle to accurately capture local details in complex images like \cref{fig:first} (a). In contrast, visual expert models excel at precise feature extraction and small object detection but lack comprehensive semantic understanding and robust multimodal integration.

\begin{figure}[t]
    \centering
    \includegraphics[width=\linewidth]{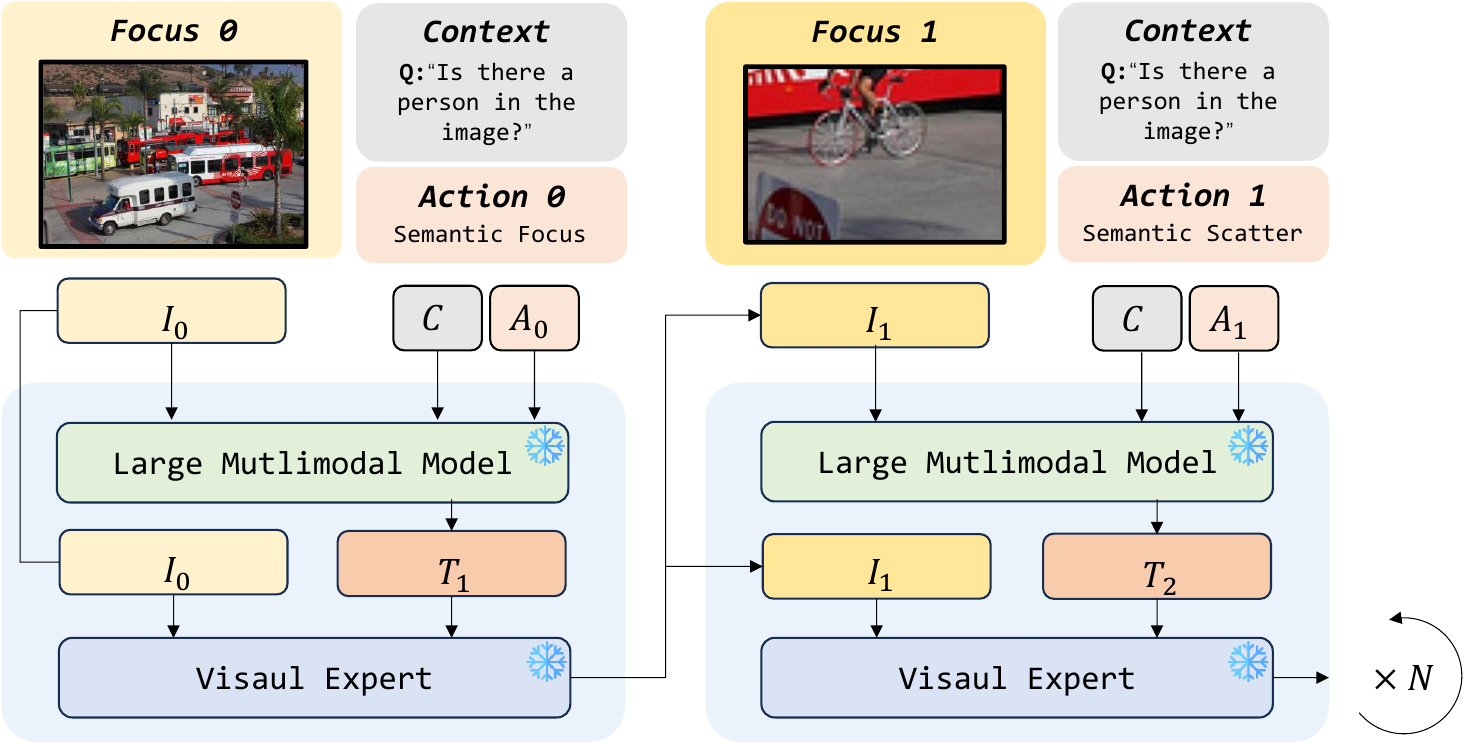}
    \caption{An illustration of Focus Adjuster of \ourmethod.}
    \label{fig:focus}
    \vspace{-12pt}
\end{figure}

To address these limitations, we propose a Focus Adjuster, a collaborative mechanism guided by action instructions that combines the strengths of both LMMs and visual experts. By achieving mutual enhancement from both textual and visual perspectives, the Focus Adjuster fuses the multimodal understanding of LMMs with the precision of visual experts, significantly improving fine-grained visual search capabilities. As shown in \cref{fig:focus}, this integration is realized in the visual search process.

We define the \focus as $f = (I, T)$, where $I \in I_o$ is the relevant image region selected from the original image $I_o$, and $T$ represents the primary semantic cue associated with $I$. The iterative update of focus $f$ proceeds as follows:
\begin{align}
\left\{
\begin{array}{l}
T^{i+1} = L(f^i, A^i), \\
I^{i+1} = E(T^{i+1}, I^i, A^i),\\
f^{i+1} = (I^{i+1},T^{i+1}),
\end{array}
\right.
\end{align}
where the action $A^i$ is an external instruction that guides focus adjustments. The LMM model $L$ refines the textual focus based on the current focus $f^i$ and action $A^i$, while the visual expert $E$ updates the image region based on the refined text cue $T^{i+1}$. This iterative process allows $A$ to dynamically guide both visual and textual update, enabling dynamic focus adjustment.

To simulate human focus-shifting actions, we designed an action space that guides the LMM and visual expert in coordinated interaction. Unlike methods that rely on LLMs to generate textual rationales~\cite{cCoT}, our actions are crafted to emulate human visual behavior rather than purely semantic reasoning.

$\diamond$ \textit{\textbf{A1}: Semantic Focus}. This action identifies contextually relevant visual targets based on semantic information (e.g., query-related objects), simulating conjunction search and endogenous orienting~\cite{berger2005infocus} to match targets according to semantic cues.

$\diamond$ \textit{\textbf{A2}: Semantic Scatter}. The action enlarges the current focused area, avoiding the loss of key information caused by overly precise focusing, simulating the divergence action of the human eye's focus.

\begin{table*}[tp]
\centering
\resizebox{0.7\linewidth}{!}{
\begin{tabular}{cllllll|l}
\hline
\textbf{Dataset}          & \textbf{Setting}                         & \textbf{Model}                & \textbf{Method} & Accuracy$\uparrow$ & Precision & Recall & F1 Score$\uparrow$  \\ \midrule
\multirow{12}{*}{MSCOCO}  & \multirow{4}{*}{\textit{Random}}      & \multirow{2}{*}{LLaVA1.5}     & Regular           &$90.07$ &$94.19$ &$85.40$ &$89.58$                                         \\
  & & & \cellcolor{blue!5}\textit{\ourmethod}               & \cellcolor{blue!5}\textbf{91.40}  & \cellcolor{blue!5}$90.86$ & \cellcolor{blue!5}$92.07$  & \cellcolor{blue!5}$\textbf{91.46}$  \\
                          &                                       & \multirow{2}{*}{Qwen2-VL} & Regular           &$90.07$ &$97.55$ &$82.20$ &$89.22$  \\
                          &                                       &                               & \textit{\ourmethod}    \cellcolor{blue!5}          &  $\textbf{92.13}$ \cellcolor{blue!5}  &  $96.95$ \cellcolor{blue!5} & \cellcolor{blue!5} $87.00$  &   $\textbf{91.71}$ \cellcolor{blue!5}  \\
                          \cline{2-8} 
                          & \multirow{4}{*}{\textit{Popular}}     & \multirow{2}{*}{LLaVA1.5}     & Regular           &$87.27$ &$88.71$ &$85.40$ &$87.02$  \\
                          &                                       &                               & \textit{\ourmethod} \cellcolor{blue!5}             & $\textbf{88.47}$ \cellcolor{blue!5}  & $85.88$ \cellcolor{blue!5} & $92.07$ \cellcolor{blue!5}  &  $\textbf{88.87}$ \cellcolor{blue!5}  \\
                          &                                       & \multirow{2}{*}{Qwen2-VL}     & Regular           & $88.57$ & $94.06$ & $82.33$ & $87.81$  \\
                          &                                       &                               & \textit{\ourmethod}   \cellcolor{blue!5}            & $\textbf{89.97}$ \cellcolor{blue!5}  & $92.13$ \cellcolor{blue!5} & $87.40$ \cellcolor{blue!5}  &  $\textbf{89.70}$ \cellcolor{blue!5}  \\
                           \cline{2-8} 
                          & \multirow{4}{*}{\textit{Adversarial}} & \multirow{2}{*}{LLaVA1.5}     & Regular           &$81.83$ &$79.86$ &$85.13$ &$82.41$  \\
                          &                                       &                               & \textit{\ourmethod}   \cellcolor{blue!5}           & $\textbf{83.40}$ \cellcolor{blue!5}  & $78.40$ \cellcolor{blue!5} & $92.20$ \cellcolor{blue!5}  &  $\textbf{84.74}$ \cellcolor{blue!5}  \\
                          &                                       & \multirow{2}{*}{Qwen2-VL}     & Regular           & $85.67$ & $88.54$ & $81.93$ & $85.11$  \\
                          &                                       &                               & \textit{\ourmethod}    \cellcolor{blue!5}           & $\textbf{86.77}$ \cellcolor{blue!5} & $86.55$ \cellcolor{blue!5} & $87.07$ \cellcolor{blue!5}  &  $\textbf{86.81}$ \cellcolor{blue!5} \\
                           \hline
\multirow{12}{*}{A-OKVQA} & \multirow{4}{*}{\textit{Random}}      & \multirow{2}{*}{LLaVA1.5}     & Regular           &$86.80$ &$82.17$ &$94.00$ &$87.69$  \\
                          &                                       &                               & \textit{\ourmethod}   \cellcolor{blue!5}            & $\textbf{88.17}$ \cellcolor{blue!5}  & $82.51$ \cellcolor{blue!5} & $96.87$ \cellcolor{blue!5}  &  $\textbf{89.11}$ \cellcolor{blue!5}  \\
                          &                                       & \multirow{2}{*}{Qwen2-VL}     & Regular           & $89.33$ & $94.49$ & $83.53$ & $88.68$  \\
                          &                                       &                               & \textit{\ourmethod}   \cellcolor{blue!5}            & $\textbf{91.40}$ \cellcolor{blue!5}  & $94.11$ \cellcolor{blue!5} & $88.33$ \cellcolor{blue!5}  &  $\textbf{91.13}$ \cellcolor{blue!5}  \\
                           \cline{2-8} 
                          & \multirow{4}{*}{\textit{Popular}}     & \multirow{2}{*}{LLaVA1.5}     & Regular           &$82.87$ &$76.88$ &$94.00$ &$84.58$  \\
                          &                                       &                               & \textit{\ourmethod}   \cellcolor{blue!5}            & $\textbf{83.70}$ \cellcolor{blue!5}  & $76.70$ \cellcolor{blue!5} & $96.80$ \cellcolor{blue!5}  &  $\textbf{85.59}$ \cellcolor{blue!5}  \\
                          &                                       & \multirow{2}{*}{Qwen2-VL}     & Regular           & $88.20$ & $92.01$ & $83.67$ &  $87.64$ \\
                          &                                       &                               & \textit{\ourmethod}   \cellcolor{blue!5}            & $\textbf{89.27}$ \cellcolor{blue!5} & $90.18$ \cellcolor{blue!5} & $88.13$ \cellcolor{blue!5}  &  $\textbf{89.14}$ \cellcolor{blue!5}  \\
                           \cline{2-8} 
                          & \multirow{4}{*}{\textit{Adversarial}} & \multirow{2}{*}{LLaVA1.5}     & Regular           &$72.57$ &$65.80$ &$94.00$ &$77.41$  \\
                          &                                       &                               & \textit{\ourmethod}     \cellcolor{blue!5}          & $\textbf{73.47}$  \cellcolor{blue!5} & $65.99$ \cellcolor{blue!5} & $96.87$ \cellcolor{blue!5}  &  $\textbf{78.50}$ \cellcolor{blue!5}  \\
                          &                                       & \multirow{2}{*}{Qwen2-VL}     & Regular           & $81.07$ & $79.61$ & $83.53$ &  $81.52$ \\
                          &                                       &                               & \textit{\ourmethod}     \cellcolor{blue!5}         & $\textbf{82.20}$ \cellcolor{blue!5}  & $78.65$ \cellcolor{blue!5} & $88.40$ \cellcolor{blue!5}  &  $\textbf{83.24}$ \cellcolor{blue!5}  \\
                           \hline
\multirow{12}{*}{GQA}     & \multirow{4}{*}{\textit{Random}}      & \multirow{2}{*}{LLaVA1.5}     & Regular           &$87.67$ &$82.66$ &$95.33$ &$88.54$  \\
                          &                                       &                               & \textit{\ourmethod}      \cellcolor{blue!5}         & $\textbf{88.50}$ \cellcolor{blue!5}  & $82.91$ \cellcolor{blue!5} & $97.00$ \cellcolor{blue!5}  &  $\textbf{89.40}$ \cellcolor{blue!5}  \\
                          &                                       & \multirow{2}{*}{Qwen2-VL}     & Regular           &$78.50$ & $98.09$ & $58.13$ & $73.00$  \\
                          &                                       &                               & \textit{\ourmethod}      \cellcolor{blue!5}         & $\textbf{86.60}$ \cellcolor{blue!5}  & $95.75$ \cellcolor{blue!5} & $76.60$ \cellcolor{blue!5}  &  $\textbf{85.11}$ \cellcolor{blue!5}  \\
                           \cline{2-8} 
                          & \multirow{4}{*}{\textit{Popular}}     & \multirow{2}{*}{LLaVA1.5}     & Regular           &$79.90$ &$72.82$ &$95.40$ &$82.60$  \\
                          &                                       &                               & \textit{\ourmethod}      \cellcolor{blue!5}         & $\textbf{80.27}$ \cellcolor{blue!5} & $72.70$ \cellcolor{blue!5} & $96.93$ \cellcolor{blue!5} &  $\textbf{83.09}$ \cellcolor{blue!5} \\
                          &                                       & \multirow{2}{*}{Qwen2-VL}     & Regular           & $78.07$ & $96.67$ & $58.13$ &  $72.61$ \\
                          &                                       &                               & \textit{\ourmethod}       \cellcolor{blue!5}        & $\textbf{84.77}$ \cellcolor{blue!5}  & $91.36$ \cellcolor{blue!5} & $76.80$ \cellcolor{blue!5} &  $\textbf{83.45}$ \cellcolor{blue!5}  \\
                           \cline{2-8} 
                          & \multirow{4}{*}{\textit{Adversarial}} & \multirow{2}{*}{LLaVA1.5}     & Regular           &$73.50$ &$66.36$ &$95.33$ &$78.25$  \\
                          &                                       &                               & \textit{\ourmethod}      \cellcolor{blue!5}         & $\textbf{74.07}$ \cellcolor{blue!5}  & $66.50$ \cellcolor{blue!5} & $97.00$ \cellcolor{blue!5}  &  $\textbf{78.90}$ \cellcolor{blue!5}  \\
                          &                                       & \multirow{2}{*}{Qwen2-VL}     & Regular           & $75.90$ & $90.09$ & $58.20$ & $70.72$ \\
                          &                                       &                               & \textit{\ourmethod}       \cellcolor{blue!5}        & $\textbf{81.97}$ \cellcolor{blue!5}  & $85.39$ \cellcolor{blue!5} & $77.13$ \cellcolor{blue!5}  &  $\textbf{81.05}$ \cellcolor{blue!5}  \\
                           \bottomrule
\end{tabular}
}
\caption{Results on POPE. \textit{Regular} denotes LMM directly answer question based on input image. \textit{\ourmethod} refers to our method. The best performances within each setting are \textbf{bolded}.}
\vspace{-1em}
\label{tab:pope}
\end{table*}

\subsubsection{Focus Tree Search} \label{sec:mcts}

We propose Focus Tree Search (FTS), a Monte Carlo Tree Search (MCTS)~\cite{kocsis2006bandit, coulom2007efficient} based visual search method that navigates through visual spaces with \focus nodes. FTS efficiently combines focus adjustment actions, balancing exploration and exploitation to construct a focus tree that retains key visual information while eliminating irrelevant details. In our approach, FTS incrementally builds a focus tree, where each node represents a \focus state \( f \), and each edge corresponds to a focus-swift action that transitions to a new focus region. To efficiently guide the Large Multimodal Model (LMM) and visual expert in exploring the most promising nodes, we maintain a state-action value function \( Q : f \times \mathcal{A} \mapsto \mathbb{R} \), which estimates the \emph{expected future reward} of performing action \( a \) in \focus state $f$ .

\vspace{1em}
\noindent \textbf{Selection Phase.}  
The algorithm starts at the root node (initial focus state \( f^0 \)) and iteratively selects the next node at each tree level by balancing exploration and exploitation. This phase continues until a leaf node is reached. To achieve the balance between exploration (visiting less-visited nodes) and exploitation (visiting high-value nodes), we use the well-established \emph{Upper Confidence bounds applied to Trees (UCT)}~\cite{kocsis2006bandit} to select each child node. Specifically, at node \( f \), we select an action by considering both the \( Q \)-value (for exploitation) and uncertainty (for exploration):
{
\small
\begin{align}
    a^\ast = \arg\max_{a \in A(f)} \left[ Q(f, a) + w \sqrt{\frac{\ln N(f)}{N(c(f, a))}} \right],
\end{align}
}

\noindent where \( N(f) \) represents the number of visits to node \( f \) in previous iterations, and \( c(f, a) \) is the child node resulting from applying action \( a \) in state \( f \). The second term increases with uncertainty (fewer visits to the child node). The weight \( w \) controls the exploration-exploitation trade-off.

\vspace{1em}
\noindent \textbf{Expansion Phase.}  
During this phase, new child nodes are added to the selected leaf node by randomly sampling unexplored actions. If the leaf node is a terminal node (i.e., the search count has reached its maximum), the expansion phase is skipped, and start backpropagation immediately.

\vspace{1em}
\noindent \textbf{Backpropagation Phase.}  
Once a terminal state is reached in the above phase, the search path from the root node to the terminal node is obtained. At this point, rewards are backpropagated along the path to update the $Q$ value for each state-action pair on the path. Specifically, \( Q(f, a) \) is updated by aggregating the rewards from all subsequent steps of node \( f \), as follows: 
{
\small
\begin{align}
Q(f, a) = \Sigma_{f^i \in N_f} R_{f^i},    
\end{align}
}

\noindent where \( N_f \) is the set of child nodes of node \( f \), and \( R_{f^i} \) is the reward value corresponding to each child node.

To encourage the LMM and visual expert to filter out irrelevant content while avoiding the introduction of biases (e.g., the visual expert locating the wrong target or the LMM generating visual hallucinations), we use consistency between the LMM and the visual expert as a criterion, with the effective area of the node serving as the search reward:
{
\small
\begin{align}
R_{f^i} = \mathbb{I}_{\{I = T\}} \cdot \frac{s_{f^i}}{s_o},    
\end{align}
}

\noindent where \( R_{f^i} \) represents the value of the reward function. \( \mathbb{I}_{\{I = T\}} \) is an indicator function that takes a value of 1 if the image \( I \) associated with node \( f^i \) is semantically consistent with the text \( T \), and 0 otherwise. \( \frac{s_{f^i}}{s_o} \) denotes the effective area ratio of node \( f^i \), where \( s_{f^i} \) is the effective area and \( s_o \) is the area of the original input image.

\begin{table*}[t]
\begin{center}
\scalebox{0.75}{
\begin{tabular}{l|l|>{\columncolor{blue!5}}c>{\columncolor{blue!5}}ccccccccc}
\toprule
\textbf{Sampling} & \textbf{Metrics} & \textit{\ourmethod-Q} & \textit{\ourmethod-L} & Osprey-7B & Ferret-7B & Shikra-7B & LLaVA-1.5& Qwen2-VL & InstructBLIP & MiniGPT4 &  mPLUG-Owl  \\
\midrule
\multirow{5}{*}{Random}&Accuracy $\uparrow$ & \textbf{92.13} & 91.40 & 89.47 & 90.24 & 86.90 & 88.73& 90.07 & 88.57 & 79.67  &   53.97 \\
& Precision & 96.95 & 90.91 & 93.40 & 97.72 & 94.40 & 88.89& 97.55 & 84.09 & 78.24 &   52.07 \\
& Recall & 87.00 & 92.00& 84.93 & 83.00 & 79.26 & 88.53& 82.20 & 95.13 & 82.20 &   99.60 \\
& F1 Score $\uparrow$ & \textbf{91.71} & 91.45 & 88.97 & 89.76 & 86.19 & 88.71& 89.22 & 89.27 & 80.17 &   68.39 \\
& Yes (\%) & 48.65 & 50.60 & 45.47 & 43.78 & 43.26 & 49.80& 43.92 & 56.57 & 52.53  &   95.63 \\
\midrule
\multirow{5}{*}{Popular}&Accuracy $\uparrow$ & \textbf{89.97} & 88.57 & 87.83 & 84.90 & 83.97 & 85.83& 88.57 & 82.77 & 69.73 &  50.90 \\
& Precision & 92.13 &86.09 & 89.94 & 88.24 & 87.55 & 83.91& 94.06 & 76.27 & 65.86 &   50.46 \\
& Recall & 87.40 & 92.00 & 85.20 & 80.53 & 79.20 & 88.67& 82.33 & 95.13 & 81.93 &   99.40 \\
& F1 Score $\uparrow$ & \textbf{89.70} & 88.95 & 87.50 & 84.21 & 83.16 & 86.22& 87.81 & 84.66 & 73.02 &   66.94 \\
& Yes (\%) & 48.59 & 53.43 & 47.37 & 45.63 & 45.23 & 52.83& 46.72 & 62.37 & 62.20 &   98.57 \\
\midrule
\multirow{5}{*}{Adversarial}&Accuracy $\uparrow$ & \textbf{86.77} & 83.43 & 85.33 & 82.36 & 83.10 & 72.10 & 85.67& 65.17 & 79.20 &   50.67 \\
& Precision & 86.55 & 78.41 & 85.43 & 83.60 & 85.60 & 74.69& 88.54 & 65.13 & 61.19 &   50.34 \\
& Recall & 87.07 & 92.27 & 85.20 & 80.53 & 79.60 & 88.34& 81.93 & 95.13 & 82.93 &   99.33 \\
& F1 Score $\uparrow$ & \textbf{86.81} & 84.78 & 85.31 & 82.00  & 82.49 & 80.94& 85.11 & 77.32 & 70.42 &   66.82 \\
& Yes (\%) & 49.65 & 58.83 & 49.87 & 48.18 & 46.50 & 59.14& 45.82 & 73.03 & 67.77 &   98.67\\
\bottomrule
\end{tabular}
}
\vspace{-2.5mm}
\caption{Results on POPE-COCO benchmark comparing our method with existing LMMs. \textit{\ourmethod-Q} is our method with Qwen2-VL, \textit{\ourmethod-L} is our method with LLaVA-v1.5. The best performances within each setting are \textbf{bolded}.}
\label{tab:pope_compare}
\end{center}
\vspace{-6.5mm}
\end{table*}

\vspace{1em}
\noindent \textbf{Multi-grained Voting}
Once the above phases are complete and the focus tree is constructed, our goal is to fully utilize the essential visual information within this tree to achieve more precise fine-grained visual understanding. Inspired by the self-consistency method~\cite{wang2023selfconsistencyimproveschainthought}, we adopt a multi-node voting approach to obtain the final answer. Specifically, the final result is obtained through a weighted voting scheme across different nodes, where each node \( f \) provides a prediction \( L(f) \), and the weight for each node is the reward \( R_f \) it has received. Formally, the final answer $\mathcal{A}$ is derived as:

{
\small
\begin{align}
\mathcal{A} = \arg\max_{L(f)} \sum_{f \in \mathcal{T}_f} R_f \cdot L(f)
\end{align}
}

\noindent where $\mathcal{T}_f$ is the focus tree composed. This voting mechanism ensures that the model avoids an overemphasis on details that could lead to the loss of critical global hint while still preserving the importance of relevant details.

\section{Experiments}
\label{sec:experiments}



\subsection{Experimental Settings}

\noindent\textbf{Datasets \& Evaluation Metrics}

\textbf{POPE} (Polling-based Object Probing Evaluation) \cite{li2023evaluating} offers a general benchmark for assessing visual hallucination in LMMs. This evaluation framework systematically tests model accuracy in identifying object presence within images by presenting balanced queries of existent and non-existent objects (50\% each). The benchmark includes three distinct sampling strategies for constructing negative samples: \textit{random}, \textit{popular}, and \textit{adversarial}. Specifically, in the \textit{random} setting, absent objects are chosen randomly; in the \textit{popular} setting, they are selected from a pool of frequently appearing objects; and in the \textit{adversarial} setting, objects are chosen based on common co-occurrences yet are absent from the image. Moreover, the POPE benchmark integrates data from three sources—MSCOCO~\cite{lin2014microsoft}, A-OKVQA~\cite{schwenk2022okvqa}, and GQA~\cite{hudson2019gqa}. Across these sampling strategies, 500 images are drawn from each dataset, with 6 queries generated per image, resulting in a total of 27,000 query-answer pairs. Finally, evaluation is conducted using four core metrics: Accuracy, Precision, Recall, and F1 score.

\textbf{V* Bench} \cite{wu2024vstar} is a challenging benchmark designed to assess large vision-language models (LMMs) on fine-grained detail recognition in high-resolution images. Based on the SA-1B dataset, it includes 191 images with an average resolution of 2246$\times$1582 and covers two key tasks: attribute recognition (115 samples) and spatial relationship reasoning (76 samples). Attribute recognition evaluates models on identifying detailed object attributes like color, material, and shape, while spatial relationship reasoning testes models on determining relative spatial positions among objects. Accuracy is used as the evaluation metric.

\begin{table}[t]
\centering
\scalebox{0.8}{
\begin{tabular}{rccc}
\toprule
& Attribute (\%) & Spatial (\%)   & Overall (\%)   \\ \midrule
Human & 98.26 & 100.00 & 98.95  \\ 
Random Guess & 26.73 & 50.00 & 35.99  \\ \midrule
\multicolumn{4}{c}{\emph{Open-source end-to-end LMMs}}  \\   
BLIP2 \cite{blip2} & 26.95 & 53.94 & 37.69   \\
MiniGPT-4 \cite{minigpt4} & 30.43 & 50.00 & 38.22  \\
LLaVA \cite{llava} & 23.47 & 53.94 &  35.59  \\
InstructBLIP \cite{instructblip} & 25.21 & 47.36 & 34.02  \\
Otter \cite{otter} & 26.95 & 56.57 & 38.74  \\ 
LLaVA-1.5 \cite{llava1.5} & 43.47 & 56.57 & 48.68 \\
Qwen2-VL \cite{Qwen2VL} & 74.78 & 68.42 & 72.25 \\
\midrule
\multicolumn{4}{c}{\emph{Commercial chatbot systems}}  \\ 
Bard \cite{bard} & 31.30 & 46.05 & 37.17   \\
Gemini Pro \cite{Gemini} & 40.86  & 59.21 & 48.16 \\
GPT-4V \cite{gpt_4}& 51.30 & 60.52 & 
 54.97 \\ \midrule
\multicolumn{4}{c}{\emph{LLM tool-using pipelines}}  \\  
MM-React \cite{yang2023mm} &34.78 &51.31  &41.36 \\
VisualChatGPT \cite{visualChatGPT} & 30.43 & 48.68 & 37.69  \\
Visprog \cite{gupta2023visual} & 31.30 & 56.57 & 41.36  \\ \midrule
\multicolumn{4}{c}{\emph{Visual search}}  \\  
SEAL \cite{wu2024vstar}  & 74.78 & 76.31 & 75.39  \\ 
 \rowcolor{blue!5} \textit{\ourmethod-L} (ours)  & 62.61   & 53.95 & 59.16 \textcolor{Green}{(+10.48)}  \\
 \rowcolor{blue!5} \textit{\ourmethod-Q} (ours)  & \textbf{80.00} & \textbf{82.89} & \textbf{81.15} \textcolor{Green}{(+8.90)} \\
\bottomrule
\end{tabular}}
\caption{Results on V* Bench comparing our method with existing LMMs. }
\vspace{-1.5em}
\label{tab:vstar}
\end{table}

\vspace{0.2cm}
\noindent\textbf{Baseline Methods} 
We assessed the effectiveness of our proposed \ourmethod on two representative LMMs: LLaVA-1.5, a fixed-resolution model, and Qwen2-VL, a variable-resolution model, using their 7B versions unless otherwise specified. To further validate the competitiveness of our approach, we compared it with recent approaches aimed at enhancing the fine-grained visual understanding capabilities of LMMs~\cite{instructblip,minigpt4, ye2023mplugowl,blip2,otter}. These include models that improve regional comprehension with additional data and training phrase \cite{Yuan2024osprey,you2023ferret,chen2023shikra}, LLMs employing multi-tool combinations to extend visual capabilities \cite{yang2023mm,visualChatGPT,gupta2023visual}, closed-source commercial chatbot systems (GPT-4V~\cite{gpt_4}, Gemini Pro~\cite{team2023gemini}, Bard\cite{bard}), and visual search methods tailored for LMMs equipped with localization and heatmap output modules \cite{wu2024vstar}.

\vspace{0.2cm}
\noindent\textbf{Implementation Details}  
In our experiments, to minimize the operational costs associated with visual experts, we utilize Lang-Segment-Anything \cite{langsam} as the sole visual expert. This system efficiently provides detection boxes and masks based on input text through batch processing, which responds faster and easy to deployment. All experiments were conducted on two NVIDIA L40 GPUs.

\begin{table}[tp]
\centering
\resizebox{0.85\linewidth}{!}{%
\begin{tabular}{@{}lcccc@{}}
\toprule
\textbf{Model}                & \textbf{$A_1$} &  \textbf{$A_2$} & Attribute(\%)$\uparrow$ & Spatial(\%)$\uparrow$ \\ \midrule
\multirow{3}{*}{LLaVA-v1.5}     & \checkmark  &         &    57.50      &  55.50           \\
                              &    &  \checkmark      &   45.46       &    52.61         \\
                              & \checkmark & \checkmark                &    62.61      &   53.95          \\ \midrule
\multirow{3}{*}{Qwen2-VL} & \checkmark    &        &    76.78      &     70.25        \\
                              &  &    \checkmark        &    75.67      &    69.50       \\
                              & \checkmark &  \checkmark   &    80.00     &     82.89                  \\\bottomrule
\end{tabular}
}
\caption{Exploration of action space combination on V* Bench.}
\label{tab:action}
\end{table}

                              

\begin{table}[tp]
\centering
\resizebox{0.85\linewidth}{!}{%
\begin{tabular}{@{}llcc@{}}
\toprule
\textbf{Model}                & \textbf{Method} & Accuracy$\uparrow$ & F1 Score$\uparrow$ \\ \midrule
\multirow{5}{*}{Random}     &  Visual Expert \cellcolor{gray!25}          &   88.87 \cellcolor{gray!25}       &  88.53 \cellcolor{gray!25}          \\
                              & LLaVA-v1.5                &  88.73        &  88.71           \\ 
                              & Qwen2-VL & 90.07 & 89.22 \\
                              &  \textit{\ourmethod-L} & 91.40 & 91.45 \\
                              &  \textit{\ourmethod-Q} & \textbf{92.13} & \textbf{91.71} \\
                              
                              \midrule
\multirow{5}{*}{Popular}     &  Visual Expert  \cellcolor{gray!25}         &   85.47 \cellcolor{gray!25}       &  85.58 \cellcolor{gray!25}          \\
                              & LLaVA-v1.5                &  85.83       &  86.22           \\ 
                              & Qwen2-VL & 88.57 & 87.81 \\
                              &  \textit{\ourmethod-L} & 88.57 & 88.95 \\
                              &  \textit{\ourmethod-Q} & \textbf{89.97} & \textbf{89.70} \\
                              \bottomrule
\end{tabular}
}
\caption{Exploration of the impact of visual expert on POPE-COCO with random and popular setting.}
\vspace{-1em}
\label{tab:expert}
\end{table}

\begin{table}[t]
\centering
\scalebox{0.8}{
\begin{tabular}{lcc}
\toprule
\textbf{Category} & \textbf{Method} & \textbf{Search Length} $\downarrow$ \\ \midrule

\multirow{2}{*}{Space Design} 
    & Uniform Dividing Policy     & 4.90 \\
    & Dynamical Focus \cellcolor{blue!5}     & \textbf{3.20} \cellcolor{blue!5} \\ \midrule

\multirow{4}{*}{Search Algorithm} 
    & A* Search                          & 4.33 \\
    & MCTS \cellcolor{blue!5}                              & \textbf{3.20} \cellcolor{blue!5}\\
    & BFS                                & 5.65 \\
    & DFS                                & 6.15 \\ 

  \bottomrule

\end{tabular}}
\caption{Comparison of search length across different search algorithms and search space design on V* Bench. The best result is in \textbf{bold}.}
\vspace{-1em}
\label{tab:speed}
\end{table}

\subsection{Experimental Results}

\noindent\textbf{POPE Benchmark Results}  
We first evaluate the potential of our method to enhance the general visual capabilities of LMMs on the POPE benchmark across three datasets (COCO, AOKVQA, GQA) and three sampling strategies (random, popular, adversarial). The results are summarized in Table~\ref{tab:pope}. A key observation is that our proposed \ourmethodlong (\ourmethod) consistently improves the performance of both fixed-resolution and dynamic-resolution models across all datasets and settings. These results suggest that \ourmethod effectively reduces visual hallucinations by emphasizing relevant visual content and minimizing the impact of irrelevant factors, thereby enhancing the visual understanding of both fixed-resolution and variable-resolution LMMs. To further assess the competitiveness of our approach, we compare it with several existing methods aimed at improving fine-grained visual understanding in LMMs. Based on the results in Table~\ref{tab:pope_compare}, \ourmethod consistently outperforms the comparison methods across all three sampling settings (random, popular, and adversarial) in the COCO dataset. These results demonstrate that \ourmethod, by combining focus tree search and expert collaboration, is able to effectively filter out irrelevant or misleading visual content, reducing hallucinations.

\vspace{0.2cm} \noindent\textbf{Results on V* Bench}
The V* Bench evaluation set focuses on fine-grained visual understanding tasks with high-resolution images and small object targets, posing challenges for LMMs in focusing details. As shown in Table~\ref{tab:vstar}, most large multimodal models (LMMs) perform near random levels, and current tool-based methods also struggle to achieve meaningful results. Notably, as a visual search method requiring no additional adaptation or training of LMMs, \ourmethod surpasses the visual search model SEAL, which fine-tunes LMMs specifically for the visual search task in both architecture and paradigm. This demonstrates the potential of flexible visual search methods that integrate visual experts with LMMs. 

\subsection{Ablation Study}
\noindent\textbf{Effect of Action Space on Visual Localization}
We further analyzed the effect of action space design  on LMM performance in fine-grained visual understanding. Table~\ref{tab:action} shows that using only single actions results in performance declines, while combining even two actions brings measurable gains.

\vspace{0.1cm} \noindent\textbf{Impact of visual expert}
To evaluate the advantages of our method compared to standalone visual experts and independent LMMs, we compared our method with only visual expert on POPE-COCO with three settings. Results in Table~\ref{tab:expert} indicate that our method consistently surpasses standalone visual expert. This suggests that combining LMMs with visual experts not only benefits from the experts' visual capabilities but also enables LMMs to mitigate biases and limitations in visual experts, resulting in mutual enhancement—consistent with our initial analysis. 

\noindent\textbf{Efficiency Study}
To further evaluate the efficiency of our search space and algorithm design, we compared them with respective counterparts using the V* Bench dataset. The search objective is to locate the node with the highest reward value within fixed max depth of 5. The search process stops either upon reaching the maximum number of iterations or identifying the target node, as shown in Table~\ref{tab:speed}. \ourmethod’s search space design requires fewer search steps than the Uniform Dividing Policy taken by SEAL~\cite{wu2024vstar}. 
The result shows that our method may avoid ambiguous search steps that uniform division method have. To assess whether Monte Carlo Tree Search (MCTS) could search efficiently by balancing exploration and exploitation, we compared it with A*, BFS, and DFS search algorithms. Experimental results indicate that MCTS achieves the highest search efficiency, which we attribute to its unique capability for reward function modeling. This feature allows it to avoid greedy traps by gradually approximating the true reward distribution. Notably, while A* theoretically guarantees the shortest path under admissible conditions, in practice, we found that finding appropriate heuristics to meet these conditions is challenging, especially when the target node is uncertain. We believe this limitation causes A* to perform similarly to greedy algorithms in our setting.

\begin{figure}[t]
    \centering
    \includegraphics[width=\linewidth]{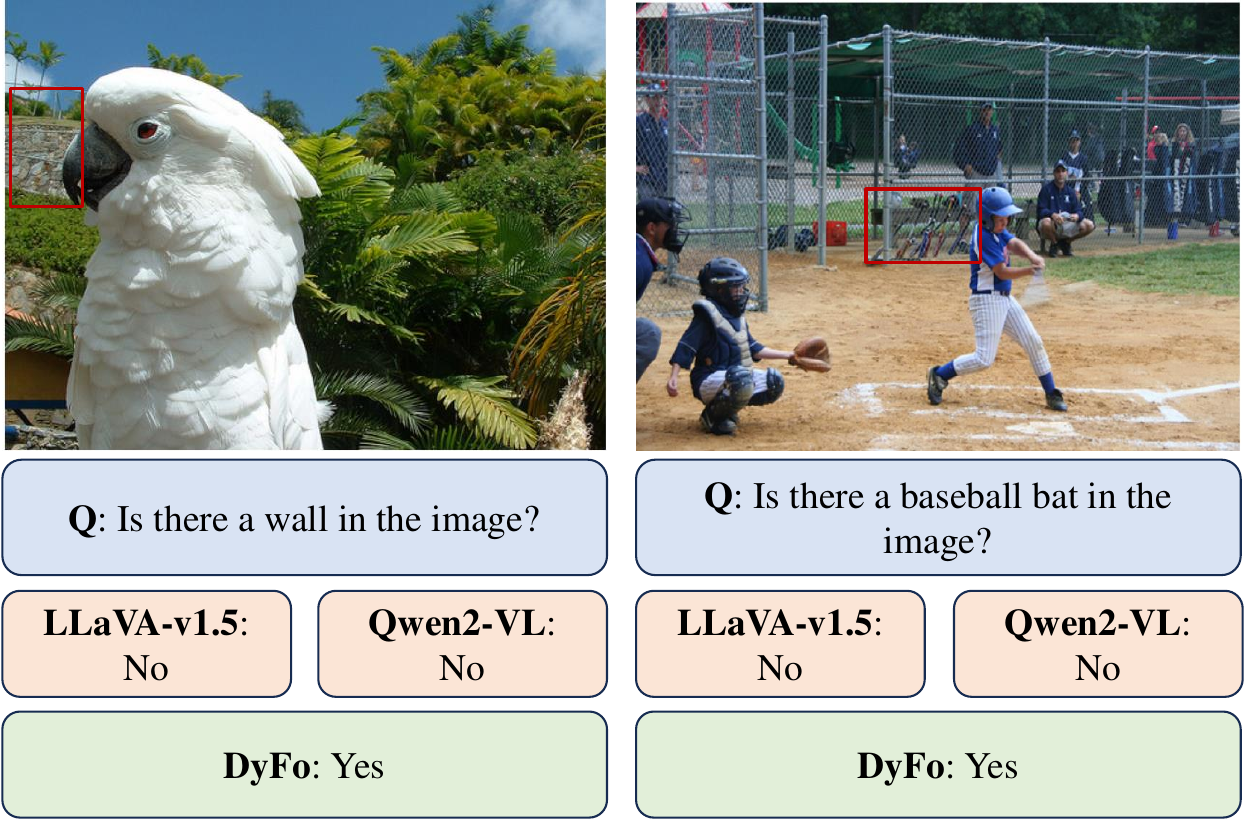}
    \caption{Comparison between the responses of LLaVA-v1.5, Qwen2-VL and our method \ourmethod on POPE cases. The final focus region is highlighted in the image using red bounding boxes.}
    \label{fig:case_pope}
\end{figure}

\begin{figure}[t]
    \centering
    \includegraphics[width=\linewidth]{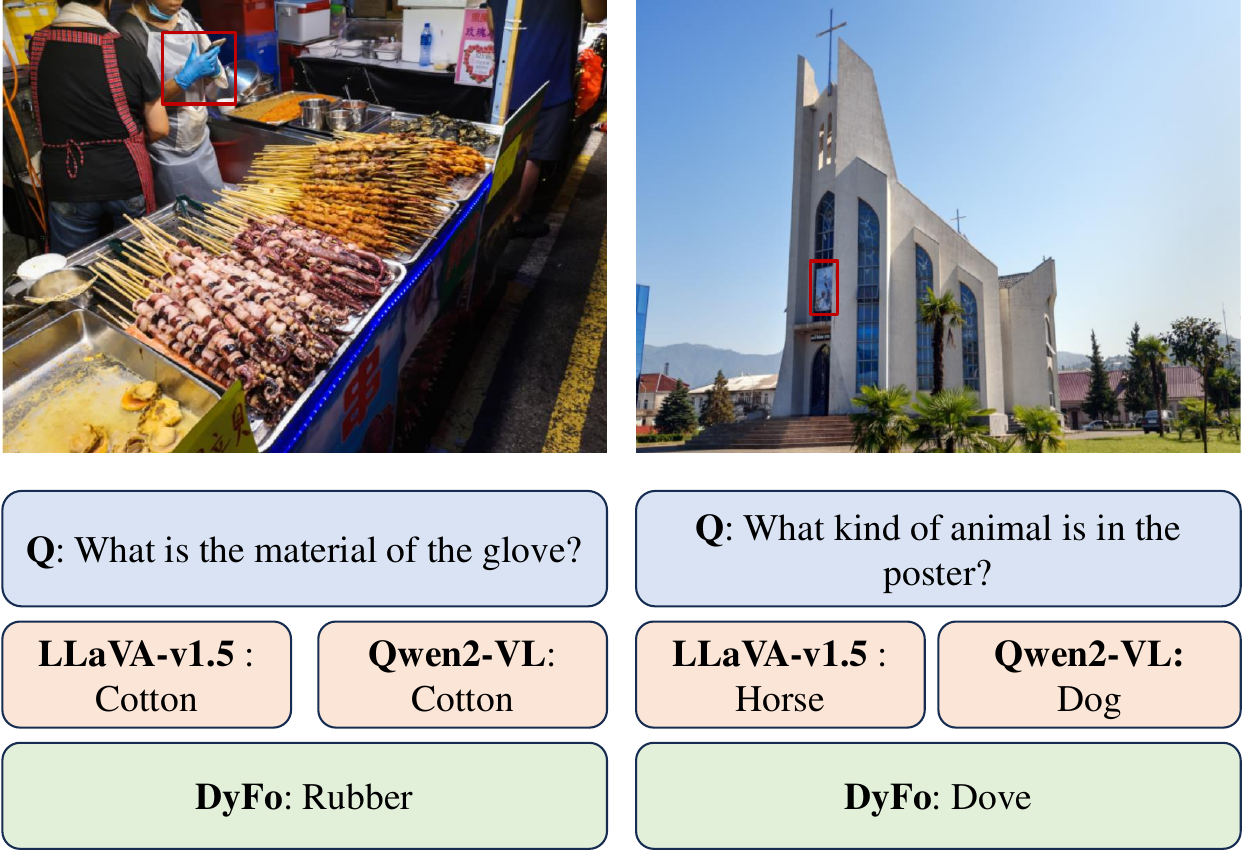}
    \caption{Comparison between the responses of LLaVA-v1.5, Qwen2-VL and our method \ourmethod on V* Bench several cases. The final focus region is highlighted in the image using red bounding boxes.}
    \label{fig:case_vstar}
    \vspace{-1em}
\end{figure}

\vspace{0.2cm}
\noindent\textbf{Case Study}
Figures~\ref{fig:case_pope}
 and \ref{fig:case_vstar} present case studies from the POPE and V* Bench datasets. In the low-resolution example in Figure~\ref{fig:case_pope}
, objects like the ``\textit{wall} " and ``\textit{baseball bat} " are off-center, partially obscured, and occupy only a small portion of the area. This makes it challenging for the LMM to accurately focus on these elements amidst a large amount of visual content, leading to hallucination errors. \ourmethod dynamically adjusts the focal area (indicated by the red box), allowing the LMM to filter out irrelevant content and mitigate inherent hallucination issues. In the high-resolution example shown in Figure~\ref{fig:case_vstar}, the challenge of focusing on key objects is even greater: objects like the ``\textit{glove}" and ``\textit{dove}" occupy less than $1/50$ of the image area, meaning that nearly 98\% of the visual input is irrelevant or distracting to the LMM. Through dynamic focusing, our method minimizes interference from irrelevant content and retains only the most critical visual information, improving performance on fine-grained visual understanding tasks.

\section{Conclusion}

We propose \ourmethod (\ourmethodlong), a training-free dynamic focus visual search method that enhances fine-grained visual understanding in large multimodal models (LMMs). By simulating human-like focusing adjustments, our method enables LMMs to selectively amplify critical visual information while filtering out irrelevant input, improving the model's ability to handle complex visual tasks with higher accuracy. Experimental results demonstrate that \ourmethod reduces visual hallucinations and enhances performance on fine-grained visual understanding tasks across various datasets suggesting that \ourmethod provides a promising approach to improving the visual capabilities of LMMs.

\section*{Acknowledgements} This work was supported by the grants from the National Natural Science Foundation of China (61925201, 62132001, 62432001, 62373043), Beijing Natural Science Foundation (L247006, 4252020) and sponsored by Tencent Basic Platform Technology Rhino-Bird Focused Research Program.%

{
    \small
    \bibliographystyle{ieeenat_fullname}
    \bibliography{main}
}


\end{document}